%% file: main.tex
\documentclass[10pt,twocolumn,letterpaper]{article}

\usepackage{cvpr}              

\usepackage{algorithmic}
\usepackage{xcolor}

\usepackage{booktabs}            

\usepackage{diagbox}             

\usepackage{multirow}            

\usepackage{makecell}
\usepackage{subcaption}
\usepackage{amsmath} 
\usepackage{tabularx} 
\usepackage{rotating}
\usepackage[utf8]{inputenc} 
\usepackage[T1]{fontenc}    

\usepackage{url}            
\usepackage{amsfonts}       
\usepackage{nicefrac}       
\usepackage{microtype}      
\usepackage{ragged2e} 
\usepackage{graphicx}
\graphicspath{{figures/}}
\usepackage{wrapfig}
\usepackage{colortbl}

\definecolor{cvprblue}{rgb}{0.21,0.49,0.74}
\usepackage[pagebackref,breaklinks,colorlinks,allcolors=cvprblue]{hyperref}
\def\paperID{12314} 
\def\confName{CVPR}
\def\confYear{2025}

\title{Binarized Mamba-Transformer for Lightweight Quad Bayer HybridEVS Demosaicing}

\author{
    Shiyang Zhou$^{1}$\thanks{Equal contribution} \hspace{0.2cm} Haijin Zeng$^{2}$\footnotemark[1]\hspace{0.2cm} Yunfan Lu$^{3}$ \hspace{0.2cm} Tong Shao$^{1}$ \\
    Ke Tang$^{1}$ \hspace{0.2cm} Yongyong Chen$^{1}$\footnotemark[2] \hspace{0.2cm} Jie Liu$^{1}$ \hspace{0.2cm} Jingyong Su$^{1}$\thanks{Corresponding author} \\
    $^{1}$Harbin Institute of Technology, Shenzhen \hspace{0.2cm} $^{2}$Harvard University \\ $^{3}$Hong Kong University of Science and Technology, Guangzhou \\
}

\begin{document}
\maketitle
\input{sec/0_abstract}

\input{sec/1_intro}
\input{sec/2_relatedwork}
\input{sec/3_methods}

\input{sec/4_experiments}
\input{sec/5_conclusion}
\input{sec/6_acknowledgement}

{
    \small
    \bibliographystyle{ieeenat_fullname}
    \bibliography{main}
}


\end{document}

%% file: sec/0_abstract.tex
\begin{abstract}

Quad Bayer demosaicing is the central challenge for enabling the widespread application of Hybrid Event-based Vision Sensors (HybridEVS). Although existing learning-based methods that leverage long-range dependency modeling have achieved promising results, their complexity severely limits deployment on mobile devices for real-world applications. To address these limitations, we propose a lightweight Mamba-based binary neural network designed for efficient and high-performing demosaicing of HybridEVS RAW images. First, to effectively capture both global and local dependencies, we introduce a hybrid Binarized Mamba-Transformer architecture that combines the strengths of the Mamba and Swin Transformer architectures. Next, to significantly reduce computational complexity, we propose a binarized Mamba (Bi-Mamba), which binarizes all projections while retaining the core Selective Scan in full precision. Bi-Mamba also incorporates additional global visual information to enhance global context and mitigate precision loss. We conduct quantitative and qualitative experiments to demonstrate the effectiveness of BMTNet in both performance and computational efficiency, providing a lightweight demosaicing solution suited for real-world edge devices. Our codes and models are available at \url{https://github.com/Clausy9/BMTNet}.
\end{abstract}

%% file: sec/1_intro.tex
\section{Introduction}
\label{sec:intro}
\begin{figure}
	\begin{center}
             \begin{subfigure}{0.48\textwidth}
                 \includegraphics[width=\linewidth]{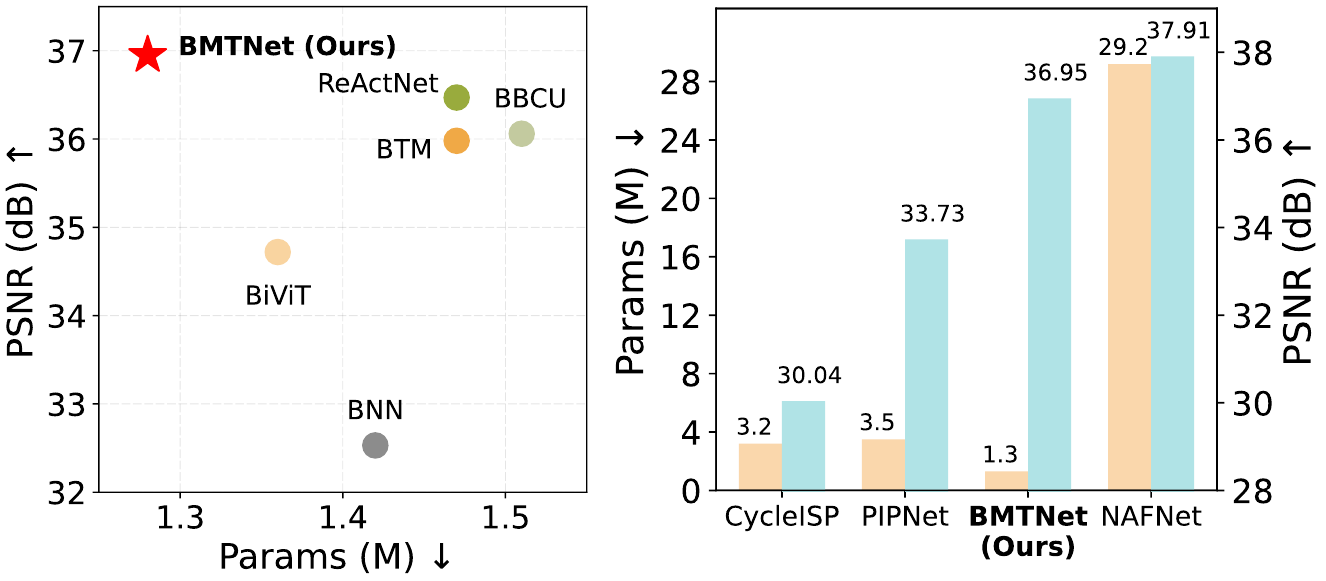}
	\vspace{-4mm}
                 \label{fig: comparison}
             \end{subfigure}
            \begin{subfigure}{0.44\textwidth}
                \begin{subfigure}{\textwidth}
                     \includegraphics[width=\linewidth]{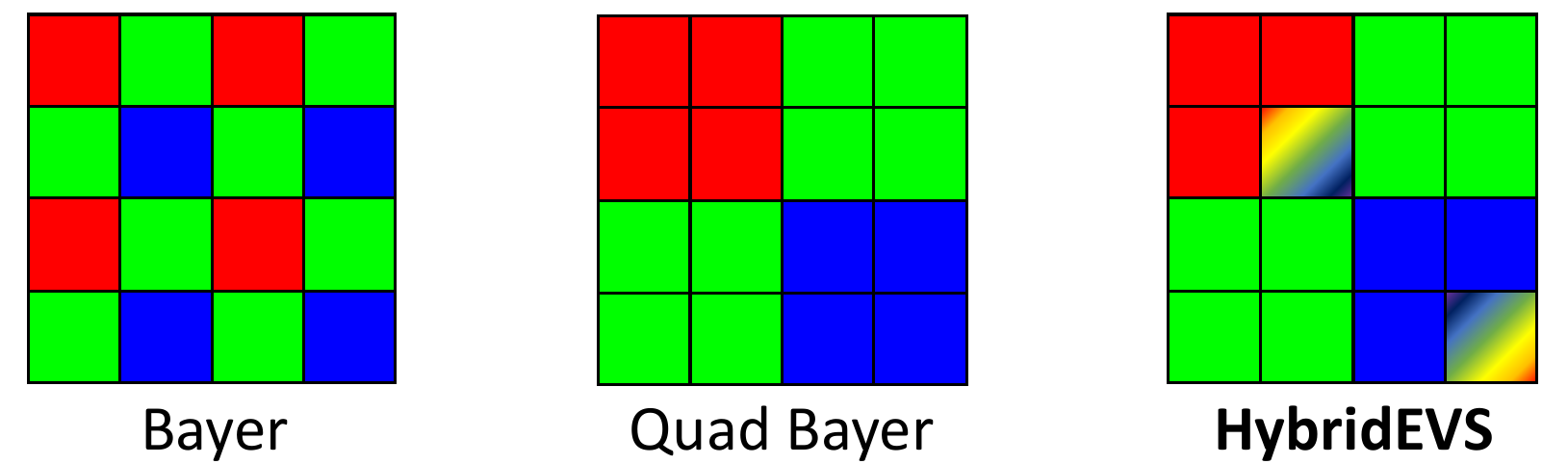}
                     \label{fig: event}
                 \end{subfigure}
             \end{subfigure}
             \hfill
	\end{center}
	\vspace{-10mm}
	\caption{\small Up-left: PSNR and parameters comparisons of our BMTNet and other BNNs on MIPI dataset. Up-right: PSNR and Parameters comparisons of our BMTNet and other FP methods on MIPI dataset. Down: CFA comparisons between Bayer, Quad Bayer, and Quad Bayer HybridEVS. Event pixel appears as a mixed color.}
	\vspace{-6mm}
	\label{figure1}
\end{figure}

By integrating traditional frame-based imaging with event-based detection, the Quad Bayer HybridEVS camera~\cite{kodama20231} has been designed as an advanced type of event camera for next-generation mobile phones. This integration captures detailed spatial information and high-speed temporal changes, offering superior performance in dynamic environments by effectively detecting motion and rapid events. Moreover, recent studies~\cite{yang2022mipi,wu2024mipi} have shown that non-Bayer color filter arrays (CFAs), like Quad Bayer, make great success in low-light scenes on mobile devices with limited sensors. These features make the Quad Bayer HybridEVS camera a cutting-edge device. However, despite its significant advantages in dynamic scenarios and low light conditions, it faces challenges in image demosaicing due to its limited sensor size, the complexity of the Quad Bayer CFA, and color loss caused by event pixels, as shown in Figure \ref{figure1}. 

Recently, with the development of mobile imaging, artificial intelligence image signal processors (AI-ISPs) have become crucial for enhancing imaging quality on mobile devices. Specifically, for the critically demosaicing process, deep learning methods provide a strong fitting ability to reconstruct full RGB images from degraded mosaic images, significantly improving restoration result 
and overcoming artifacts and aliasing issues on Quad Bayer CFA sensors~\cite{a2021beyond,zheng2024quad}. Although existing methods have demonstrated the effectiveness of deep learning in demosaicing, their deployment on edge devices remains challenging due to the high computational cost. Moreover, recent studies have shown that global information, such as long-range dependencies and global context~\cite{wu2024seesr,zhang2024recognize} plays a key role in enhancing image restoration tasks. 
These tasks are typically based on Transformer architectures, which can effectively capture global dependencies. However, these methods suffer from high complexity and insufficient emphasis on global visual priors, making it challenging for demosaicing approaches to achieve high performance with limited resources.

To address this issue, various model compression methods have been proposed, such as 8-bit and 4-bit quantization~\cite{banner2019post, banner2018scalable}. Among these, binary neural networks (BNNs) represent the most extreme approach, compressing the model to only positive and negative values, which marks the upper limit of model compression~\cite{hubara2016binarized, rastegari2016xnor, cai2024binarized}.
These highly compressed methods demonstrate significant potential for deploying deep learning models onto resource-constrained devices. Prior works have shown BNN's capability in both natural language processing and vision tasks, including large language models~\cite{huang2024billm,yuanpb}, image classification~\cite{hubara2016binarized,rastegari2016xnor,qin2020forward}, and image restoration~\cite{zhang2024binarized,cai2024binarized}, etc. However, the application of binary networks for the demosaicing task remains an unexplored area for further research.

To fully leverage global information with fewer resources, state space models (SSMs)~\cite{gu2021efficiently,smith2022simplified} have emerged as a fundamental architecture, competing with conventional structures like convolutional neural networks (CNNs) and Transformers. Advanced SSMs like Mamba~\cite{gu2023Mamba} model have demonstrated significant capability in capturing long-range dependencies with lower computational costs than self-attention mechanisms. Nevertheless, deploying Mamba on resource-limited devices is still constrained by complex projection layers. BNN offers a promising optimization to reduce complexity in less critical layers of Mamba, yet most research on binarization mainly focused on CNNs and Transformers, leaving Mamba binarization relatively blank.

In this paper, we introduce a binarized Mamba-Transformer architecture called BMTNet to tackle the aforementioned issues, as shown in Figure \ref{figure2}. The basic unit called the Binarized Mamba-Transformer (BMT) block combines Mamba and Transformer in a binary format with linear complexity, which is pioneering the binarization of Mamba by binarizing all projections, which effectively reduces model complexity while maintaining core functions in full precision. The advanced combination effectively leverages both long and short dependencies while significantly decreasing parameters and computation costs by 97\% and 96\% on Mamba blocks. 

To reduce precision loss and enhance global feature extraction, we incorporate an additional binarized global visual encoder specifically designed for Quad Bayer RAW images to capture global visual information. For the visual representation, we use an embedding mechanism that combines image features with global information to generate a control matrix for the input in the Selective Scan process of our proposed Binarized Mamba (Bi-Mamba). This property can effectively enhance the perceptual ability of BMTNet by improving the control matrix of input. Our proposed BMTNet outperforms other BNN methods and achieves results comparable to full precision methods, as shown in Figure \ref{figure1},
In summary, the contributions of our work are that
\begin{itemize}
\item We introduce a novel BNN-based hybrid BMTNet for Quad Bayer HybridEVS demosaicing. To the best of our knowledge, this is the \textbf{first} research to explore binary Mamba and binarized Quad Bayer demosaicing.
\item We propose a hybrid binary Mamba-Transformer design that benefits from its dual-branch form and linear complexity in a binary format, efficiently capturing global and local dependencies with minimal computation.  
\item An advanced binary Mamba that binarizes non-critical projections while keeping the core SSM in full precision, enabling meaningful global visual information to be embedded in the attention process and significantly reducing complexity by 79\% on parameters and 88\% on OPs.

\end{itemize}

%% file: sec/2_relatedwork.tex
\section{Related Work}
\label{sec:relatedwork}
Demosaicing is a crucial step in the imaging process, aiming to reconstruct RGB images from RAW data, and is typically integrated into ISPs. Traditional demosaicing methods are mainly based on interpolation~\cite{hirakawa2005adaptive,li2008image}. In recent years, Convolution Neural Networks (CNNs) have led a great improvement in Bayer demosaicing, exhibiting impressive ability to overcome the color degradation~\cite{tan2017color,tan2017joint,syu2018learning,tan2018deepdemosaicking,liu2020joint}. Despite demosaicing on Bayer images having achieved great success, non-Bayer CFAs, which have more complicated color arrangements present new challenges. With the development of mobile imaging, Quad Bayer has been widely implemented on flagship smartphone cameras~\cite{yang2022mipi} and advanced event cameras~\cite{son2017640}, which demonstrate better imaging results on dark scenes. Still, as an emergent CFA, research on Quad Bayer is limited. Some research~\cite {jia2022learning,zeng2023inheriting} proposed two-stage strategies on remosaicing Quad Bayer to Bayer. Some GAN-based structures~\cite{a2021beyond,sharif2021sagan} have been proposed to achieve refined detail on Quad Bayer demosaicing. However, most of them neglect constrained computation resources on ISPs, making them difficult to deploy on edge devices.

\begin{figure*}
     \centering
     \includegraphics[width=\textwidth]{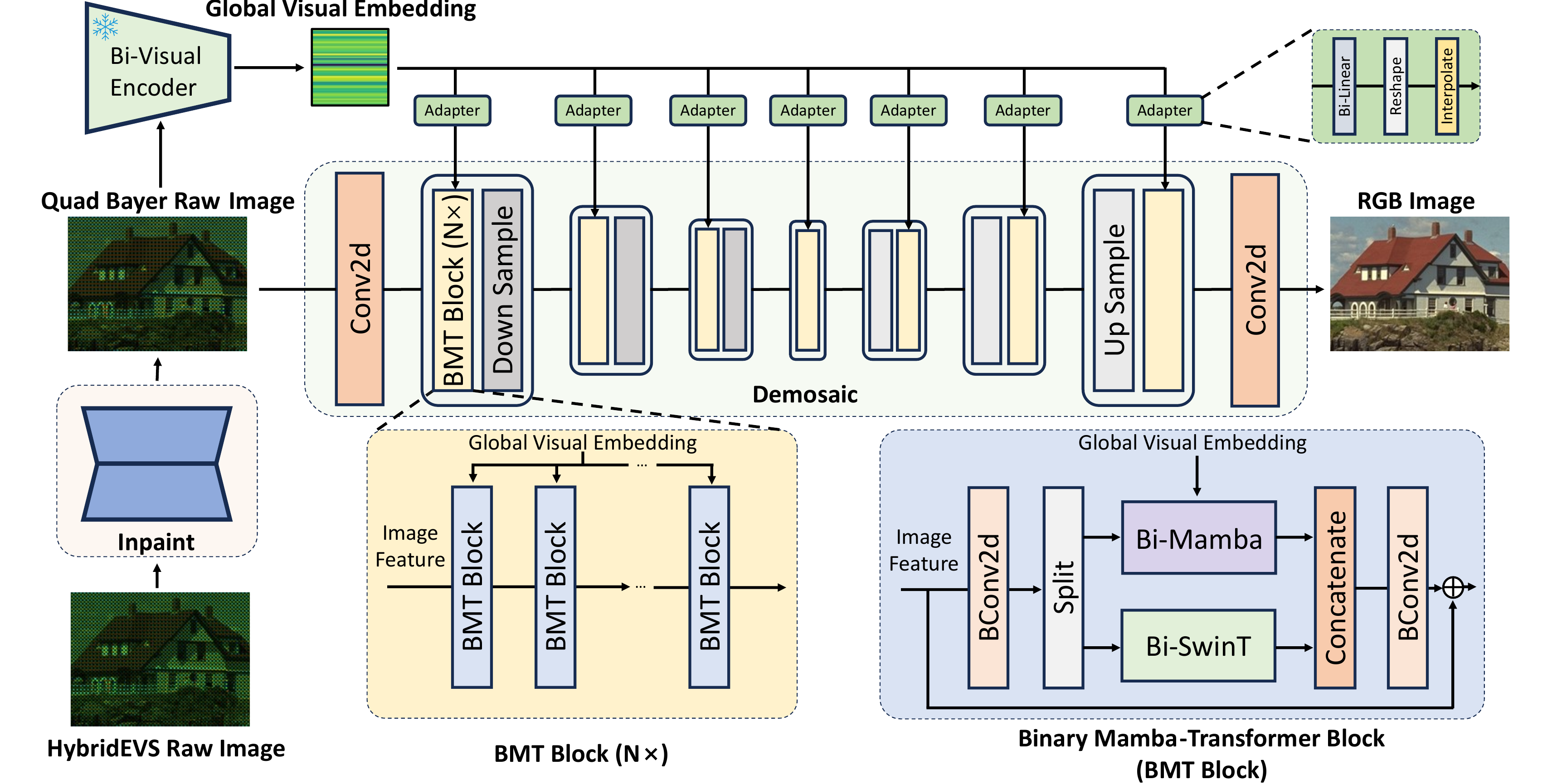}
     \vspace{-6mm}
     \caption{Overall architecture of BMTNet. A binary convolution-based simple subnetwork is initially employed for event pixel inpainting. The main branch incorporates our hybrid binary Mamba-Transformer Block, which pioneeringly integrates Bi-Mamba with Bi-Swin Transformer to capture both global and local features. An additional global visual branch is used to enhance global dependencies, with Bi-Mamba specifically handling the fusion of global features.}
    \vspace{-4mm}
    \label{figure2}
\label{fig: BMTNet}
\end{figure*}
Recently, event cameras have emerged as a bio-inspired vision sensor to capture changes in the scene, showing high dynamic range, high temporal resolution, and low latency~\cite{gallego2020event,son2017640,dong2024joint}. Some  methods~\cite{zhang2022spiking,munda2018real} demonstrate advanced applications of event cameras. Pioneering methods~\cite{ma2024color4e} explore demosaicing for neuromorphic event sensors with real data. Nevertheless, research on the recently proposed Quad Bayer Hybrid Event Vision Sensor (HybridEVS) is still limited~\cite{kodama20231}. MIPI workshop~\cite{wu2024mipi} proposed some advanced methods for demosaicing. Some advanced methods~\cite{lu2024event,xu2024demosaicformer,yunfanrgb} are proposed to solve the cutting-edge problem. However, the substantial computational demands make these methods challenging to deploy on resource-constrained ISPs. The need for the exploration of lightweight demosaicing methods remains essential.

State space models (SSMs) have developed to be a powerful competitor to CNNs and Transformers that capture global information with linear complexity. Prior models~\cite{gu2021efficiently,smith2022simplified} introduce efficient parallel scanning to improve the effectiveness and speed of SSMs. The emergence of Mamba~\cite{gu2023Mamba} features a data-independent SSM layer that demonstrated impressive performance with linear complexity. Prior works have utilized Mamba on multiple computer vision tasks~\cite{zhu2024vision,liao2024lightm,wang2023selective,hatamizadeh2024mambavision} as well as image restoration~\cite{guo2024Mambair,xiao2024frequency}. The linear scalability of complexity and strong modeling ability for long-dependencies demonstrate extraordinary potential for visual tasks, with emerging OEM prototypes exploring Mamba's edge deployment.

Binary neural network (BNN) is an extreme compression technology that quantizes the network's weights and activation values on a 1-bit form, which can bring a significant reduction in computation loads. Early works~\cite{courbariaux2015binaryconnect,hubara2016binarized,qin2020forward,liu2018bi} introduced binarized CNNs, utilizing the sign function to binarize weights and activations. To mitigate the loss of precision due to binarization, Rastegari \emph{et al.}~\cite{rastegari2016xnor} adapted scaling factors for weights and activations. For BNNs in image restoration tasks, Xia \emph{et al.}~\cite{xia2022basic} determined some essential components of BNN in image restoration. Some advanced research~\cite{cai2024binarized,zhang2024binarized} employed BNNs for downstream restoration tasks. However, the binarized network for the demosaicing task still needs to be explored. Furthermore, besides binarization of CNNs, binarized Transformers~ are also explored~\cite{qin2022bibert,liu2022bit}, He \emph{et al.}~\cite{he2023bivit} expanded binarization into vision Transformers. \emph{Still, as for the recently proposed Mamba network, no work has yet explored its binarization.}

%% file: sec/3_methods.tex
\section{Methods}
\label{sec:methods}
\begin{figure*}
     \centering
     \includegraphics[width=0.96\textwidth]{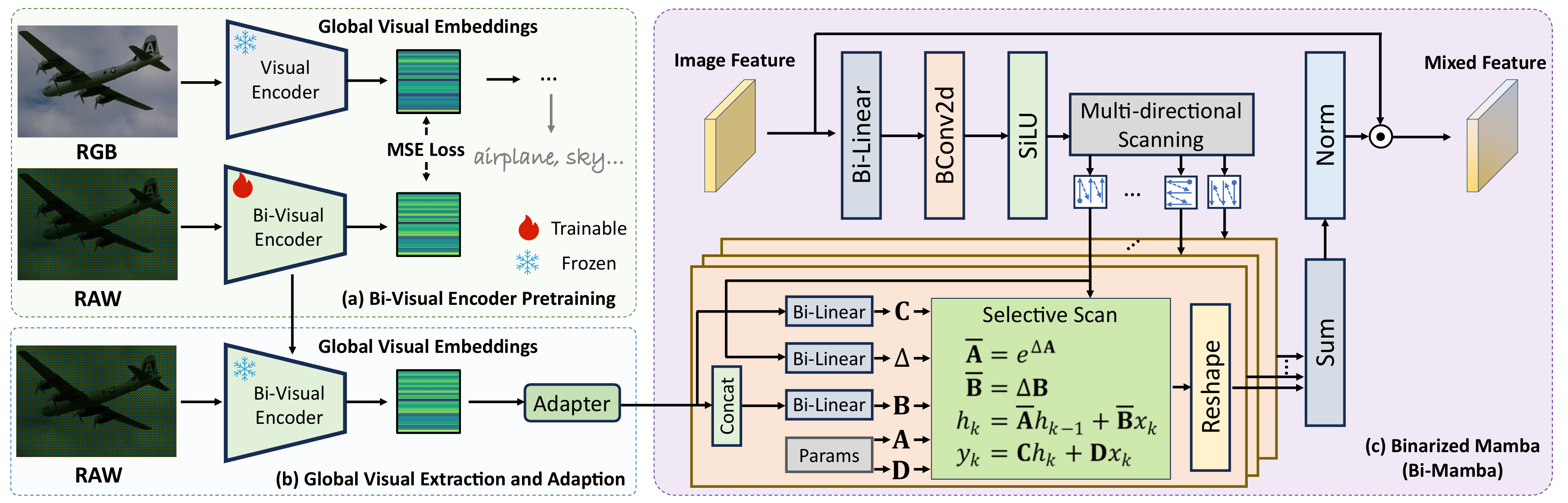}
     \vspace{-3mm}
     \caption{Model details of the bi-visual encoder and Bi-Mamba. (a) We first adopted a pretrained large visual encoder from RAM~\cite{zhang2024recognize} to pretrain our binarized visual encoder fit for Quad Bayer RAW input. (b) During the training of BMTNet, the binarized visual encoder is frozen and produces global visual embeddings to Bi-Mamba after an adapter. (c) In the binarized Mamba, we binarize all projections while keeping the core selective scan calculation in full precision, effectively reducing computational load while maintaining performance. To further enhance the global capacity, we introduce extra global information into the control matrix $\mathbf{B}$ of input.}
    \vspace{-4mm}
    \label{figure3}
\label{fig: biMamba}
\end{figure*}

In this section, we first formulate the Quad Bayer HybridEVS demosaicing problem and outline the overall architecture of our network. We then provide a detailed explanation of the proposed binarized visual encoder and binarized Mamba-Transformer block.

\subsection{Network Architecture}
Demosaicing for Quad Bayer HybridEVS cameras is a cutting-edge challenge, particularly for real-world event camera applications. Unlike standard demosaicing tasks, the unique design of HybridEVS introduces the Quad Bayer CFA and the event pixel. The Quad Bayer CFA increases color distortion due to larger gaps between identical colors, while the event pixel leads to additional color loss, making demosaicing especially challenging. The task is to reconstruct the RGB image $\mathbf{I_R} \in \mathbb{R}^{H\times W\times 3}$ from the Quad Bayer RAW image $\mathbf{I_Q} \in \mathbb{R}^{H\times W\times 1}$. Additionally, HybridEVS RAW images experience color loss at event position $\mathbf{L} \in \mathbb{R}^{H\times W\times 1}$. The relationship between these factors can be expressed as:
\begin{equation}
\mathbf{I_Q} = \mathcal{M}(\mathbf{I_R})+\mathbf{L},
\end{equation}
where $\mathcal{M}$ indicates Quad Bayer mosaic process.
To address these challenges, we introduce the binarized hybrid Mamba-Transformer network (BMTNet), as shown in Figure \ref{fig: BMTNet}. Our approach starts with a subnetwork $\mathcal{N}_{1}$ based on BBCU~\cite{xia2022basic} to coarsely inpaint color loss. Then, the binarized Mamba-Transformer network $\mathcal{N}_{2}$ solves the demosaicing task. Additionally, we incorporate an encoder branch to leverage additional global visual information from Quad Bayer images. The entire process can be formulated as:
\begin{equation}
\mathbf{I_R} = \mathcal{N}_{2}(\mathcal{N}_{1}(\mathbf{I_Q})).
\end{equation}

Next, we present the binarized visual encoder and details of the hybrid binarized Mamba-Transformer block.

\subsection{Binarized Visual Encoder} \label{SE}

Recent research shows that global visual information supplies extra global information, improving the accuracy of image restoration, especially in finer details~\cite{wu2024seesr}. To strengthen Mamba’s global capacity, we incorporated a global visual encoder tailored for Quad Bayer RAW images in a binary format, as shown in Figure \ref{fig: BMTNet}, which offers implicit visual encoding with little computational load. This encoder starts with a pretraining phase (see Figure \ref{fig: biMamba} (a)), where a frozen large visual encoder (from RAM~\cite{zhang2024recognize}) serves as a teacher to train a compact visual encoder. This enables a direct fusion of global visual information into the main branch in vector form.

During main branch training and inference, the binarized visual encoder remains frozen to preserve its capacity for global visual extraction (see Figure \ref{fig: biMamba} (b)). The visual embeddings are then fed into each layer's Bi-Mamba module through a single-layer Bi-Linear adapter, as shown in Figure \ref{fig: BMTNet}. This adapter is essential for adapting the visual embeddings across layers with different levels of information. Overall, the additional visual encoder ensures global consistency throughout the encoding and decoding process, providing a distinct prior and preserving global structural information throughout these stages.

\subsection{Binarized Mamba-Transformer}\label{bmtb}
Mamba is an efficient mechanism for capturing long-range dependencies with linear complexity, but the numerous projection layers still make it challenging to deploy on resource-constrained mobile devices. The Swin Transformer has demonstrated strong capabilities in extracting local features but is also limited by high computational demands. To leverage the strengths of both architectures while expanding the applicability of BNN, we introduce a binarized Mamba-Transformer block (BMT block), as shown in Figure \ref{figure2}. The BMT block employs a two-branch design combining the binarized Mamba (Bi-Mamba) and binarized Swin Transformer (Bi-SwinT) to capture global and local dependencies in parallel. Additionally, it benefits from a reduced channel number in each sub-block, lowering the computation load and resulting in a more lightweight model.

Specifically, the Bi-Mamba fully binarizes non-critical projections while using full precise computations for the core selective scan (SS) function, along with an extra global visual embedding on the control matrix of input, the illustration of Bi-Mamba is shown in Figure \ref{figure3} (c). This approach minimizes model complexity while maintaining high performance with precise core computations. The quantization in Mamba mainly focuses on linear and convolution projections. In our approach, full precision linear weights $\mathbf{W}^f \in \mathbb{R}^{C_{in}\times C_{out}}$ are binarized using the Sign function and activation $\mathbf{A}^f \in \mathbb{R}^{H\times W\times C_{in}}$ are binarized using RSign function~\cite{liu2020reactnet}, resulting in values of \{+1, -1\}, which can be expressed as:
\vspace{-1mm}
\begin{equation}
\mathbf{W}^{b}=\operatorname{Sign}\left(\mathbf{W}^f\right)=\left\{\begin{array}{ll}
+1, & \mathbf{W}^f>0 \\
-1, & \mathbf{W}^f \leq 0
\end{array}.\right.
\end{equation}
\vspace{-1mm}
\begin{equation}
\mathbf{A}^{b}=\operatorname{RSign}\left(\mathbf{A}^f\right)=\left\{\begin{array}{ll}
+1, & \mathbf{A}^f>\alpha \\
-1, & \mathbf{A}^f \leq \alpha
\end{array},\right.
\end{equation}
where $\alpha \in \mathbb{R}^{C_{in}}$ represents learnable parameters. To mitigate precision loss due to binarization, we apply additional learnable scaling factors $\mathbf{S} \in \mathbb{R}^{C_{out}}$ ~\cite{he2023bivit}. The whole Bi-Linear process can be expressed as: 
\begin{equation}
\begin{aligned}
\operatorname{Bi-Linear}(A^f) &= \mathbf{W}^b \ast \mathbf{A}^b \ast \mathbf{S} \\
&= \operatorname{bitcount}(\mathbf{XNOR}(\mathbf{W}^b, \mathbf{A}^b)) \ast \mathbf{S}.
\end{aligned}
\end{equation}
The Binarized convolution ($\operatorname{BConv2d}$) applies similar binarization on weights and activations but sets $\mathbf{S}$ as the average of $\mathbf{W}^f$. Specific to Bi-Mamba, given a full-precision input $\mathbf{X}^f \in \mathbb{R}^{H\times W\times C_{in}}$, it is first projected into feature $\mathbf{X'} \in \mathbb{R}^{H\times W\times d}$, where $d$ is the hidden dimension, which can be expressed as:
\begin{equation}
\mathbf{X'} =\operatorname{SiLU}(\operatorname{BConv2d}(\operatorname{Bi-Linear}(\mathbf{X}^f))).
\end{equation}
Features are then transformed into a one-dimensional format using multiple scan orders~\cite{liu2024vMambavisualstatespace}, which can be expressed as:
\begin{equation}
\begin{split}
    \mathbf{X}_1, \mathbf{X}_2, \ldots, \mathbf{X}_n =
    \operatorname{S_1}(\mathbf{X'}), \operatorname{S_2}(\mathbf{X'}), \ldots, \operatorname{S_n}(\mathbf{X'}),
\end{split}
\end{equation}
where $\operatorname{S}$ denotes the scan operation and $n$ is the total number of scan types. For each $\mathbf{X}_i \in \mathbb{R}^{L\times d}$, $i\in \{1, 2, \ldots, n\}$, $L=H*W$, we extract SSM parameters $\mathbf{B}_i \in \mathbb{R}^{L\times m}$, $\mathbf{C}_i \in \mathbb{R}^{L\times m}$ and $\mathbf{\Delta}_i \in \mathbb{R}^{L\times d}$. Unlike previous Mamba methods that derive all parameters solely from the input $\mathbf{X}_i$, we enhance the control matrix by concatenating the additional global visual vector $\mathbf{S}\in \mathbb{R}^{L\times 1}$ from the bi-visual encoder. The projection of SSM parameters is thus represented as:
\begin{equation}
    \begin{split}
 \mathbf{C}_i &= \operatorname{Bi-Linear}(\mathbf{X}_i),\\
 \mathbf{\Delta}_i &= \operatorname{Bi-Linear}(\mathbf{X}_i),\\
\mathbf{B}_i &= \operatorname{Bi-Linear}(\operatorname{Concat}(\mathbf{X}_i, \mathbf{S})).
\end{split}
\end{equation}
With the learnable parameters $\mathbf{A}_i \in \mathbb{R}^{d\times m}$ and $\mathbf{D}_i \in \mathbb{R}^{d}$, the Selective Scan (SS) process in full precise can be expressed as:
\begin{equation}\label{eq:9}
\begin{split}
\overline{\mathbf{B}}_i &= \mathbf{B}_i^\top \otimes \mathbf{\Delta}_i, \\
\overline{\mathbf{A}}_i &= \operatorname{exp}(\mathbf{\Delta}_i \otimes \mathbf{A}_i), \\
\mathbf{O}_i &= \operatorname{Reshape}(\operatorname{SSM}(\overline{\mathbf{A}}_i, \overline{\mathbf{B}}_i, \mathbf{C}_i, \mathbf{D}_i, \mathbf{X}_i)).
\end{split}
\end{equation}

This SSM~\cite{gu2023Mamba} function computes the most critical long-range attention dependencies. Retaining it in full precision enables our Bi-Mamba to achieve results comparable to full-precision Mamba with minimal computation cost. The global visual embedding in the control matrix $\mathbf{B}$ directly enriches the input $\mathbf{X}$ with extra global visual features, contributing to the recursive SS process in an additive manner. This approach preserves the original formula while effectively enhancing performance without significant disruption. The final output is computed by summing the SS results and performing a Hadamard product with the output from another Bi-Linear branch, which can be expressed as:
\begin{equation}
\mathbf{X}^{out} =
\operatorname{LN}(\sum_{i=1}^{n}\mathbf{O}_i) \ast \operatorname{SiLU}(\operatorname{Bi-Linear}(\mathbf{X}^f)),
\end{equation}
where $\operatorname{LN}$ represents layer normalization. Bi-Mamba reduces parameters by binarizing non-essential projections like linear and convolution while maintaining high performance through full-precision selective scanning and global visual embedding. This approach offers a practical way to enhance Mamba and significantly compress the model into a binary format, making it well-suited for demosaicing on resource-limited edge devices.

To further enhance local representation, we introduce the binarized Swin Transformer (Bi-SwinT) block. The Bi-SwinT block employs a binary format~\cite{he2023bivit} of the Swin Transformer~\cite{liu2021swin}, facilitating information exchange and improving local detail extraction. In summary, the integration of the binarized Mamba-Transformer block pioneeringly simplifies the Mamba architecture while effectively capturing both global and local dependencies with minimal computational loads, which significantly broadens the application of BNNs in vision Transformers and Mamba.

%% file: sec/4_experiments.tex
\begin{figure*}[ht]
    \centering
    \begin{subfigure}[b]{0.38\textwidth}
        \centering
        \includegraphics[width=\textwidth, trim=200 200 150 25, clip]{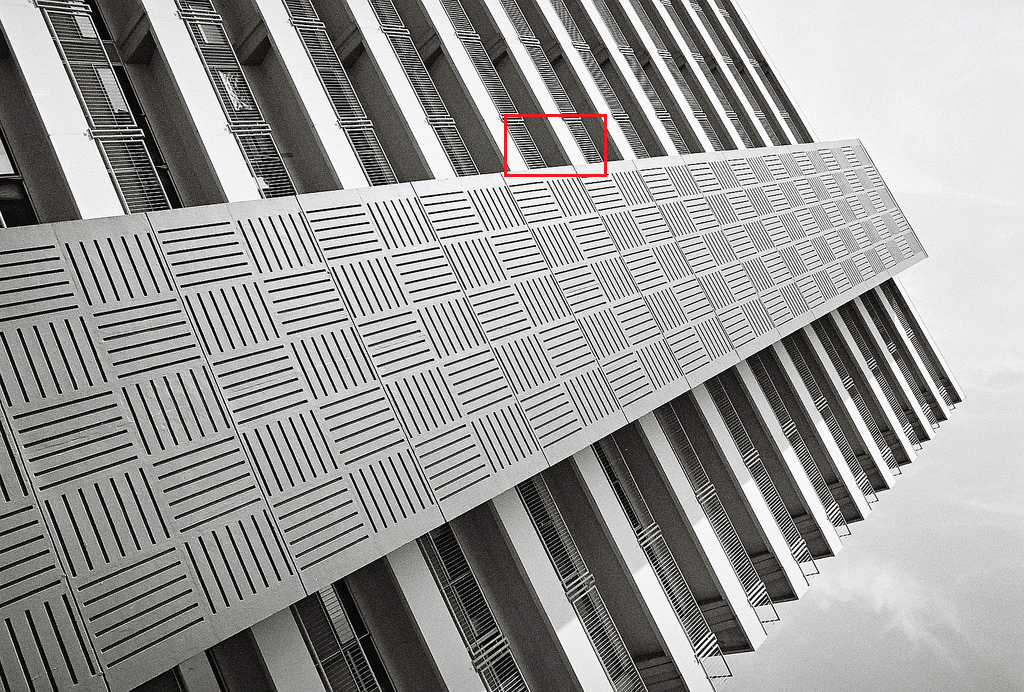}
         \caption*{\makecell{067, Urban100~\cite{Cordts_2016_CVPR}\\PSNR/SSIM}}
    \end{subfigure}%
     \hfill
    \begin{subfigure}{0.2\textwidth}
        \begin{subfigure}[b]{\textwidth}
            \centering
            \includegraphics[width=\textwidth]{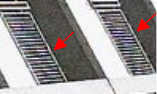}
            \caption*{\makecell{BBCU~\cite{xia2022basic}\\26.75/0.958}}
        \end{subfigure}
        \begin{subfigure}[b]{\textwidth}
            \centering
            \includegraphics[width=\textwidth]{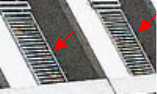}
            \caption*{\makecell{ReActNet~\cite{liu2020reactnet}\\26.04/0.961}}
        \end{subfigure}
        
    \end{subfigure}
    \begin{subfigure}{0.2\textwidth}
        \begin{subfigure}[b]{\textwidth}
            \centering
            \includegraphics[width=\textwidth]{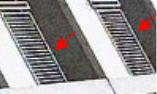}
            \caption*{\makecell{BiViT~\cite{he2023bivit}\\27.32/0.961}}
        \end{subfigure}
        \begin{subfigure}[b]{\textwidth}
            \centering
            \includegraphics[width=\textwidth]{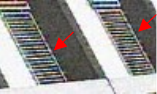}
            \caption*{\makecell{BNN~\cite{hubara2016binarized}\\22.31/0.894}}
        \end{subfigure}
        
    \end{subfigure}
    \begin{subfigure}{0.2\textwidth}
        \begin{subfigure}[b]{\textwidth}
            \centering
            \includegraphics[width=\textwidth]{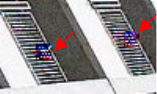}
            \caption*{\makecell{BTM~\cite{jiang2021training}\\21.05/0.929}}
        \end{subfigure}
        \begin{subfigure}[b]{\textwidth}
            \centering
            \includegraphics[width=\textwidth]{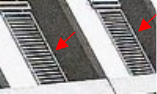}
            \caption*{\makecell{\textbf{BMTNet (Ours)}\\ \textbf{29.07/0.972}}}
        \end{subfigure}
        
    \end{subfigure}

    \vspace{-3mm}
    \caption{Visualization on the Urban100 dataset across all compared BNN methods. The proposed BMTNet achieves the best visual quality, effectively reducing artifacts and color aliasing.}
  \label{figure4}
  \vspace{-2mm}
\end{figure*}
\section{Experiments}
\begin{table*}[ht]
  
  \footnotesize
  \centering
  \fontsize{9}{11}\selectfont
  \setlength{\tabcolsep}{1.4mm}\resizebox{1\linewidth}{!}{ 
    \begin{tabular}{p{2.4cm}|cc|cccccc|c}
    \toprule
    \fontsize{10}{12}\selectfont{\multirow{2}{*}{Methods}} & Params   & \fontsize{10}{12}\selectfont{OPs} &  \multirow{2}{*}{MIPI} & \multirow{2}{*}{Kodak} & \multirow{2}{*}{McM} & \multirow{2}{*}{BSD100}& \multirow{2}{*}{Urban100} & \multirow{2}{*}{Wed} & \fontsize{10}{12}\selectfont{\multirow{2}{*}{Average}}
    \\    
    & (M) & (G)&&&&&&&
    \\
    \midrule
\rowcolor{blue!2}
DFormer~\cite{Xu_2024_CVPR} & 30.28 & 491.1 & \textbf{39.35}/0.981 & \textbf{39.32}/0.982 & \textbf{37.88}/\textbf{0.963} & \textbf{37.65}/0.982 & \textbf{37.64}/0.980 & \textbf{34.86}/0.968 & \textbf{38.15}/0.978 \\
\rowcolor{blue!2}
NAFNet~\cite{chen2022simple} & 29.16 & 32.19 & 37.91/0.980 & 38.60/0.984 & 36.18/0.961 & 37.12/0.985 & 35.63/0.978 & 35.24/\textbf{0.972} & 37.16/0.978 \\
\rowcolor{blue!2}
Restormer~\cite{zamir2022restormer} & 26.11 & 282.2 & 38.46/\textbf{0.984} & 39.16/\textbf{0.986} & 36.54/\textbf{0.967} & 37.11/\textbf{0.985} & 36.36/0.977 & 35.00/0.971 & 37.45/\textbf{0.980} \\
\rowcolor{blue!2}
SAGAN~\cite{sharif2021sagan} & 22.56 & 341.6 & 34.25/0.959 & 36.14/0.974 & 32.58/0.939 & 30.53/0.931 & 29.89/0.946 & 28.22/0.917 & 32.74/0.952 \\
\rowcolor{blue!2}
PIPNet~\cite{a2021beyond} & 3.46 & 68.8 & 33.73/0.950 & 32.20/0.960 & 31.34/0.928 & 31.97/0.950 & 28.92/0.942 & 29.19/0.929 & 31.97/0.951 \\
\rowcolor{blue!2}
CycleISP~\cite{zamir2020cycleisp} & 3.23 & 104.9 & 30.04/0.934 & 33.09/0.970 & 30.37/0.919 & 32.18/0.969 & 29.78/0.942 & 30.22/0.944 & 31.14/0.952 \\
\midrule
\rowcolor{green!5}
BNN~\cite{hubara2016binarized} & 1.42 & 6.45 & 32.53/0.930 & 33.83/0.955 & 28.59/0.889 & 31.43/0.953 & 29.56/0.935 & 29.32/0.926 & 31.07/0.926 \\
\rowcolor{green!5}
ReActNet~\cite{liu2020reactnet} & 1.47 & 6.12 & 36.47/0.971 & 37.25/0.978 & 34.55/0.946 & 35.25/0.978 & 33.86/0.970 & 33.53/0.962 & 35.08/0.970 \\
\rowcolor{green!5}
BBCU~\cite{xia2022basic} & 1.51 & 6.97 & 36.06/0.970 & 37.03/0.978 & 33.74/0.941 & 34.30/0.977 & 33.27/0.967 & 32.63/0.959 & 34.85/0.967 \\
\rowcolor{green!5}
BTM~\cite{jiang2021training} & 1.47 & \textbf{6.12} & 35.98/0.972 & 37.39/0.979 & 32.99/0.947 & 35.41/0.979 & 33.69/0.971 & 32.76/0.962 & 34.59/0.972 \\
\rowcolor{green!5}
BiViT~\cite{he2023bivit} & 1.36 & 6.51 & 34.72/0.963 & 36.55/0.975 & 30.44/0.932 & 33.48/0.974 & 32.79/0.965 & 30.40/0.952 & 33.33/0.963 \\
\rowcolor{gray!20}
\textbf{BMTNet (Ours)} & \textbf{1.28} & 6.56 & \textbf{36.95}/\textbf{0.975} & \textbf{37.69}/\textbf{0.980} & \textbf{34.79}/\textbf{0.950} & \textbf{36.11}/\textbf{0.981} & \textbf{34.45}/\textbf{0.973} & \textbf{33.95}/\textbf{0.965} & \textbf{35.52}/\textbf{0.975} \\
    \bottomrule
\end{tabular}}
  \vspace{-2mm}
  \caption{Quantitative evaluation of our BMTNet compared to full-precision and other BNN methods across six image datasets, using PSNR (dB) / SSIM as evaluation metrics for visual quality. The blue background indicates full-precise methods, while the green background means binary neural networks. BMTNet outperforms other BNNs while achieving results comparable to full-precision models at a minimal computational cost.}
  \label{tab:table1}
  \vspace{-5mm}
\end{table*}
\label{sec:experiments}
In this section, we first specify the implementation details. Then, we evaluate our BMTNet across eight diverse datasets, including both single images and video sequences, demonstrating a comparison with full precision and BNN methods. Finally, we perform an ablation study to analyze our methods. Additional analysis of extended visual results is provided in the supplementary material.

\subsection{Experimental Settings}
\paragraph{Datasets.}
The dataset used in our experiments comprises both simulated and real data. The simulated data includes ground truth, which can be utilized for training and quantitative analysis, while the real data lacks definitive values and is employed for qualitative analysis.
We train all models using the MIPI dataset from the CVPR Mobile Intelligent Photography \& Imaging Workshop 2024. This dataset includes 800 training pairs and 26 test pairs of RAW and RGB images, containing synthesized Gaussian noise and defect pixels~\cite{wu2024mipi}. To broaden the test scenes in different conditions, we simulate the HybridEVS test cases on additional seven datasets, including image datasets: Kodak~\cite{10.1145/1290082.1290117}, McM~\cite{Woo_2018_ECCV}, BSD100~\cite{937655}, Urban100~\cite{Cordts_2016_CVPR}, Wed~\cite{7752930}, and video datasets: REDS~\cite{nah2019ntire}, Vid4~\cite{liu2011bayesian}. These datasets cover a wide range of daily life scenes. 
The \textbf{real-world data} was collected using a hybrid-vision sensor (HVS) developed in collaboration with Alpsentek~\cite{alpsentek2024}, which is based on the ALPIX-Eiger chip, with a resolution of $2448 \times 3246$. Additionally, we collected scenes in both indoor and outdoor environments. 
For the demosaicing task, we specifically captured a 2000-line resolution chart to evaluate the resolution performance of different methods. 
To focus the evaluation on demosaicing, we employed long exposure to minimize the noise in the RAW images and applied white balance after processing the results to prevent color deviation.
Furthermore, we used MATLAB to perform a basic ISP on the RAW images to obtain a reference image for comparison.
\begin{figure*}[!ht]
     \begin{subfigure}[b]{0.135\textwidth}
             \centering
             \includegraphics[width=\linewidth]{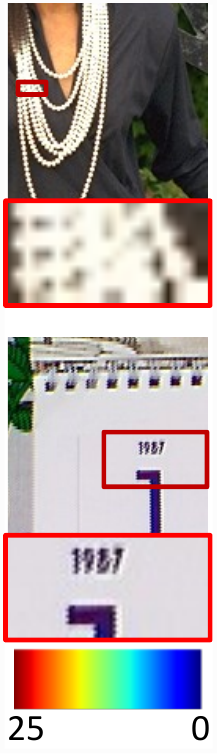}
         \caption{Reference}
     \end{subfigure}
     \hfill
     \begin{subfigure}[b]{0.135\textwidth}
         
             \centering
             \includegraphics[width=\linewidth]{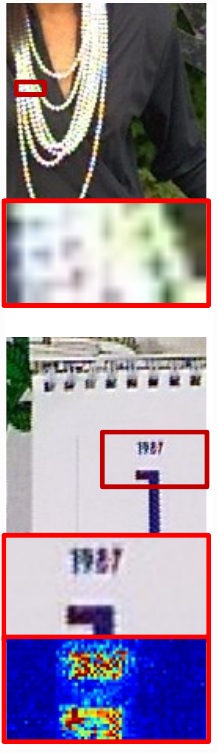}
         
         \caption{BNN~\cite{hubara2016binarized}}
        \end{subfigure}
     \hfill
     \begin{subfigure}[b]{0.135\textwidth}
         
             \centering
             \includegraphics[width=\linewidth]{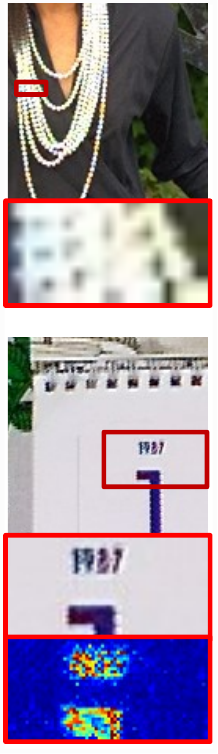}
         
         \caption{BBCU~\cite{xia2022basic}}
         \end{subfigure}
     \hfill
     \begin{subfigure}[b]{0.135\textwidth}
         
             \centering
             \includegraphics[width=\linewidth]{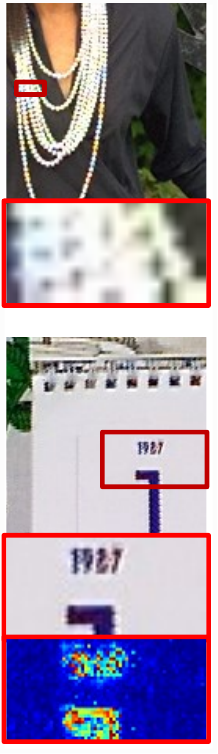}
         \caption{BiViT~\cite{he2023bivit}}
         \end{subfigure}
     \hfill
     \begin{subfigure}[b]{0.135\textwidth}
         
             \centering
             \includegraphics[width=\linewidth]{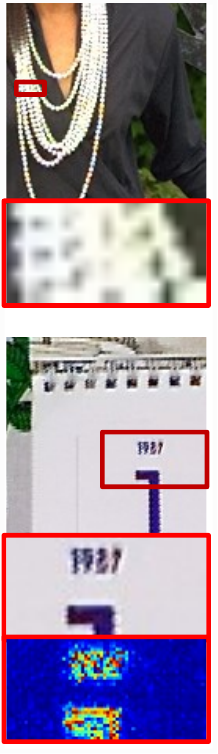}
         \caption{BTM~\cite{jiang2021training}}
         \end{subfigure}
     \hfill
     \begin{subfigure}[b]{0.135\textwidth}
         
             \centering
             \includegraphics[width=\linewidth]{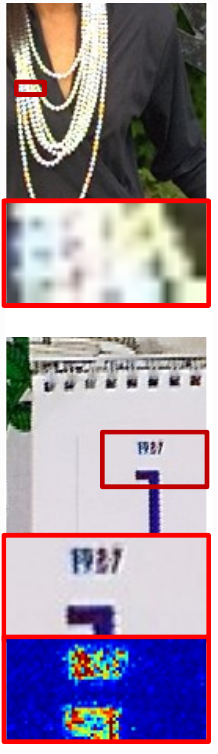}
         \caption{ReActNet~\cite{liu2020reactnet}}
         \end{subfigure}
     \hfill
     \begin{subfigure}[b]{0.135\textwidth}
             \centering
             \includegraphics[width=\linewidth]{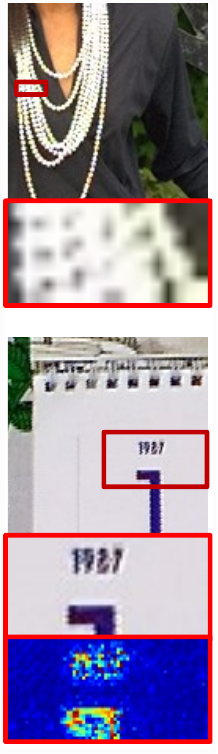}
         \caption{\textbf{BMTNet (Ours)}}
     \end{subfigure}
     \vspace{-2mm}
    \caption{Visualized results across all compared BNN methods on the Kodak (up) and Vid4 (down) datasets, with a corresponding heatmap showing the pixel value differences. The proposed BMTNet exhibits less color aliasing than other BNN methods.}
    \label{figure5}
    \vspace{-3mm}
\end{figure*}

\vspace{-3mm}
\paragraph{Implementation Details.} 
During training, we randomly crop images into 128 $\times$ 128 patches with batch size = 32. The Adam optimizer with $L1$ loss is employed, with a learning rate from $2 \times 10^{-4}$ to $1 \times 10^{-7}$ in a cosine annealing scheme. Total iterations are set to $1 \times 10^{6}$. For BMTNet and other compared BNNs, we apply a pretraining step to utilize the two-stage structure. To preserve performance, upsampling and downsampling operations remain in FP.

\vspace{-3mm}
\paragraph{Computation Load Calculation of BNNs.} Following the prior works on BNN~\cite{hubara2016binarized,liu2018bi,xia2022basic}, binarized operations ($\operatorname{BOPs}^b$) is computed as $\operatorname{BOPs}^b = \operatorname{BOPs}^f/64$, with total operations calculated by $\operatorname{OPs} = \operatorname{BOPs}^b + \operatorname{OPs}^f$, where $\operatorname{OPs}^f$ means floating-point operations. The binarized parameters are calculated as  $\operatorname{BParms}^b = \operatorname{BParms}^f/32$, and total parameters are computed as  $\operatorname{Parms} = \operatorname{BParms}^b + \operatorname{Parms}^f$, where $\operatorname{Parms}^f$ indicates the number of float-point parameters. All computational load tests are conducted on a 256$\times$256 image.
\begin{table}[ht]
  
  \footnotesize
  \centering
  \fontsize{9}{11}\selectfont
  \setlength{\tabcolsep}{1.2mm}{
    \begin{tabular}{p{2.4cm}|cc|c}
    \toprule
    
    \fontsize{9}{11}Methods  &REDS& Vid4& Avearge\\
    
    \midrule
\rowcolor{blue!2}
    DFormer~\cite{Xu_2024_CVPR} & 42.45/0.991 & 36.01/0.979 & 39.23/0.985 \\
\rowcolor{blue!2}
    NAFNet~\cite{chen2022simple} & 41.65/0.989 & 34.98/0.974 & 38.31/0.982 \\
\rowcolor{blue!2}
    Restormer~\cite{zamir2022restormer} & 41.91/0.990 & 35.08/0.981 & 38.50/0.985 \\
\rowcolor{blue!2}
    SAGAN~\cite{sharif2021sagan} & 38.13/0.984 & 32.16/0.963 & 35.15/0.974 \\
\rowcolor{blue!2}
    PIPNet~\cite{a2021beyond} & 36.19/0.981 & 32.20/0.964 & 34.20/0.973 \\
\rowcolor{blue!2}
    CycleISP~\cite{zamir2020cycleisp} & 32.96/0.975 & 30.46/0.964 & 31.71/0.965 \\
\rowcolor{blue!2}
    \midrule
\rowcolor{green!5}
    BNN~\cite{hubara2016binarized} & 36.55/0.978 & 31.23/0.957 & 33.89/0.967 \\
\rowcolor{green!5}
    ReActNet~\cite{liu2020reactnet} & 40.28/0.991 & 33.90/0.975 & 37.09/0.983 \\
\rowcolor{green!5}
    BBCU~\cite{xia2022basic} & 40.47/0.992 & 33.79/0.974 & 37.13/0.983 \\
\rowcolor{green!5}
    BTM~\cite{jiang2021training} & 40.30/0.992 & 33.35/0.975 & 36.83/0.983 \\
\rowcolor{green!5}
    BiViT~\cite{he2023bivit} & 39.96/0.990 & 33.59/0.972 & 36.78/0.981 \\
    \rowcolor{gray!20}
    \textbf{BMTNet (Ours)} & \textbf{41.15/0.993} & \textbf{34.24/0.976} & \textbf{37.70/0.985} \\
    
    \bottomrule
  \end{tabular}
  
  }
  \vspace{-2mm}
  \caption{Quantitative evaluation on two video datasets, using PSNR (dB) / SSIM as evaluation metrics. BMTNet outperforms all other BNNs by over 0.6dB in PSNR, surpasses several full-precision networks, and approaches state-of-the-art FP methods.}
  \label{tab:table2}
  \vspace{-5mm}
\end{table}

\vspace{-1mm}
\subsection{Comparison to State-of-the-Arts}
\vspace{-1mm}
\paragraph{Quantitative Comparison.}
We compared our BMTNet with other binarization methods by replacing BMT block to the corresponding BNN block, including BNN~\cite{hubara2016binarized}, ReActNet~\cite{liu2020reactnet}, BBCU~\cite{xia2022basic}, BTM~\cite{jiang2021training} and  BiViT~\cite{he2023bivit}. Additionally, we compare our BMTNet with State-of-the-Arts image restoration and demosaicing networks, including DFormer~\cite{Xu_2024_CVPR}, NAFNet~\cite{chen2022simple}, Restormer~\cite{zamir2022restormer}, SAGAN~\cite{sharif2021sagan}, CycleISP~\cite{zamir2020cycleisp}, PIPNet~\cite{a2021beyond}. 
\begin{figure*}[ht]
    \centering
    \begin{subfigure}[b]{0.135\textwidth}
             \centering
             \includegraphics[width=\linewidth]{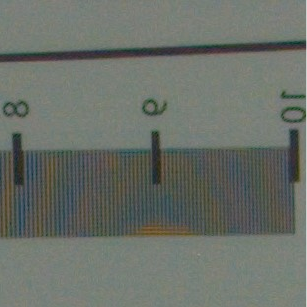}
         \caption{Reference}
     \end{subfigure}
     \hfill
     \begin{subfigure}[b]{0.135\textwidth}
         
             \centering
             \includegraphics[width=\linewidth]{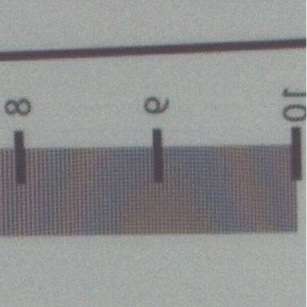}
         
         \caption{BNN~\cite{hubara2016binarized}}
        \end{subfigure}
     \hfill
     \begin{subfigure}[b]{0.135\textwidth}
         
             \centering
             \includegraphics[width=\linewidth]{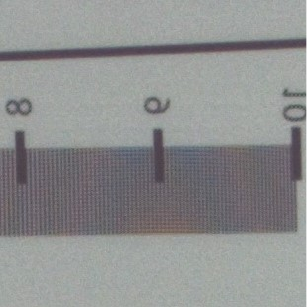}
         
         \caption{BBCU~\cite{xia2022basic}}
         \end{subfigure}
     \hfill
     \begin{subfigure}[b]{0.135\textwidth}
         
             \centering
             \includegraphics[width=\linewidth]{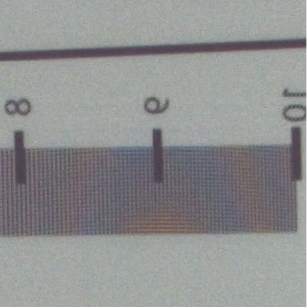}
         \caption{BiViT~\cite{he2023bivit}}
         \end{subfigure}
     \hfill
     \begin{subfigure}[b]{0.135\textwidth}
         
             \centering
             \includegraphics[width=\linewidth]{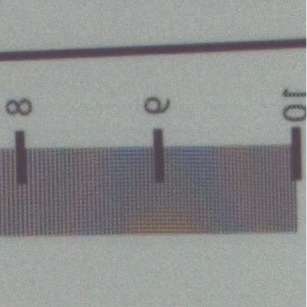}
         \caption{BTM~\cite{jiang2021training}}
         \end{subfigure}
     \hfill
     \begin{subfigure}[b]{0.135\textwidth}
         
             \centering
             \includegraphics[width=\linewidth]{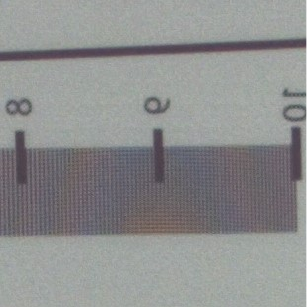}
         \caption{ReActNet~\cite{liu2020reactnet}}
         \end{subfigure}
     \hfill
     \begin{subfigure}[b]{0.135\textwidth}
             \centering
             \includegraphics[width=\linewidth]{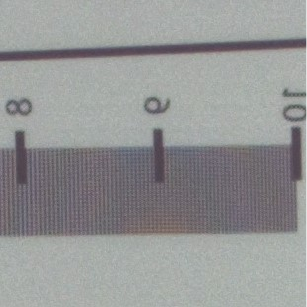}
         \caption{\textbf{BMTNet (Ours)}}
     \end{subfigure}

    \vspace{-1mm}
    \caption{Visualized comparison on real data of HybridEVS. Reference is acquired from the classic demosaic method. Our BMTNet reduces moiré artifacts on dense lines and achieves the best result among BNNs.}
  \label{figure6}
  \vspace{-2mm}
\end{figure*}
We evaluated the performance and computational load of the models across six image datasets and two video datasets, as illustrated in Table \ref{tab:table1} and Table \ref{tab:table2}. The upper section reports the performance of full precision models, while the lower section demonstrates the results of BNNs. Notably, some full-precision methods perform unsatisfied results due to the color loss caused by event pixels. In contrast, our BMTNet demonstrates robust performance with significant reductions in parameters and operations. Furthermore, our proposed binary Mamba-Transformer block outperformed other BNN methods on all image and video datasets, achieving superior results with the least parameters of 1.28M and a slight increase of operators of 0.44G compared with the minimized model BTM due to the full-precise Selective Scan. The results validate that our method effectively enhances information extraction across both local and global dimensions.

\vspace{-4mm}
\paragraph{Visual Comparison.}
Visual comparisons are presented in Figures \ref{figure4}, \ref{figure5}, and \ref{figure6}, showing results on image data, video data, and real HybridEVS images. Our BMTNet reaches superior visual performance, which effectively reduces color aliasing and moiré artifacts in the test datasets compared to other approaches. Our BMTNet also demonstrates the best visual results on real-world HybridEVS data, effectively reducing artifacts when facing dense lines on images.

\begin{figure}
	\begin{center}
            \begin{subfigure}{0.235\textwidth}
                 \includegraphics[width=\linewidth]{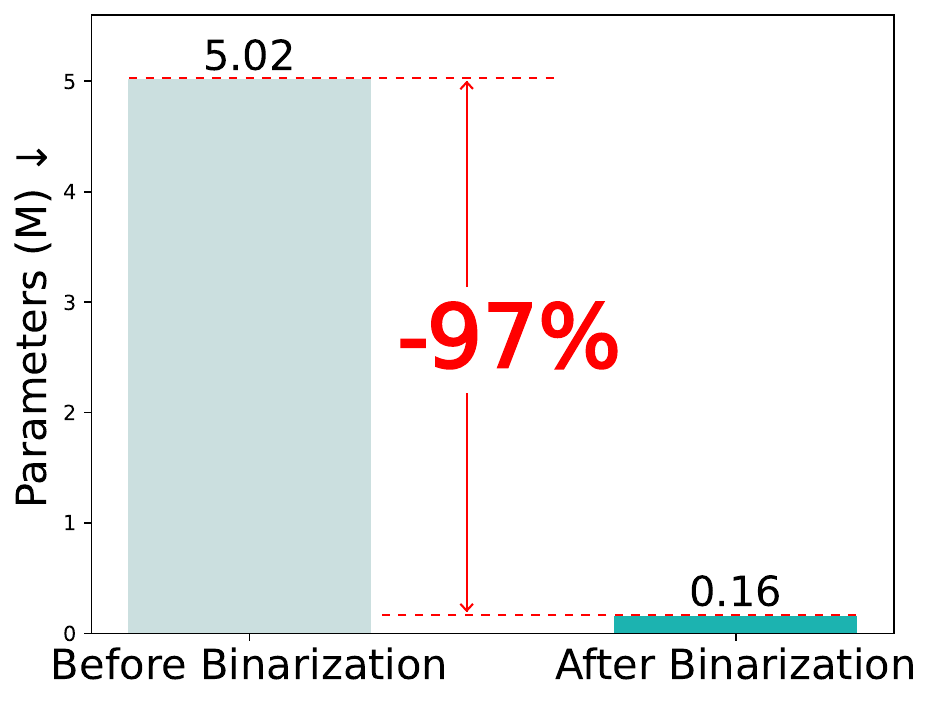}
             \end{subfigure}
             \hfill
             \begin{subfigure}{0.235\textwidth}
                 \includegraphics[width=\linewidth]{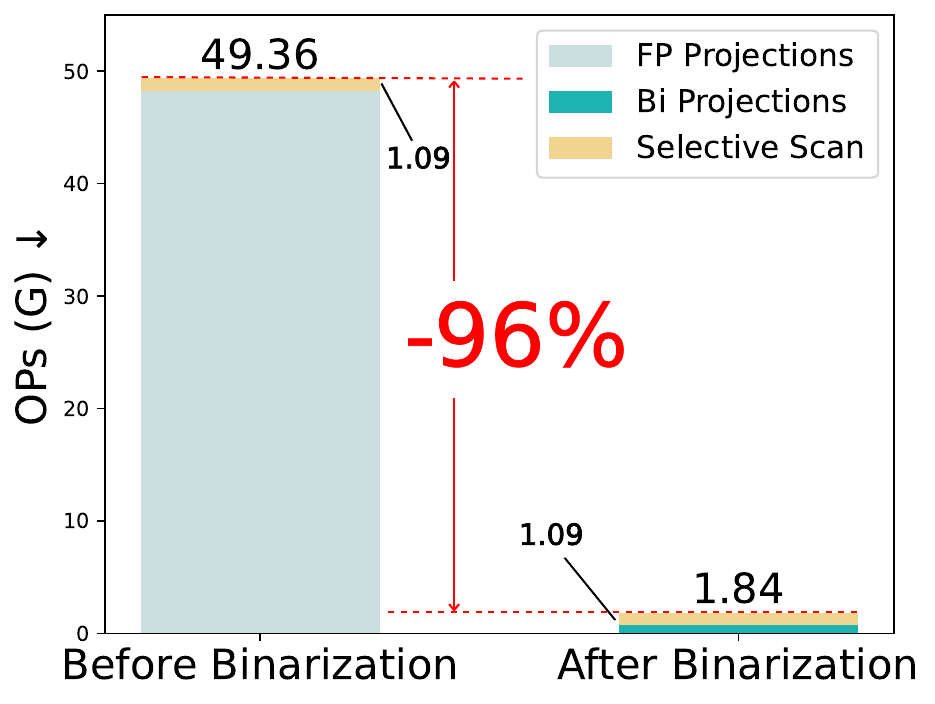}
             \end{subfigure}
	\end{center}
	\vspace{-5mm}
	\caption{\small Left: Parameters reduction of our Bi-Mamba. Right: Operations reduction of our Bi-Mamba.}
	\vspace{-4mm}
	\label{figure7}
\end{figure}

\subsection{Ablation Study}
We demonstrate an ablation study on the proposed binarized Mamba block to validate its effectiveness, as shown in Table \ref{tab:table3} and Table \ref{tab:table4}, separately conducting the influence of binarization and global visual embeddings. 

\vspace{-4mm}
\paragraph{Binarization of Mamba}
As shown in Table \ref{tab:table3}, our proposed Bi-Mamba achieves a significant PSNR improvement of 1.78dB, attributed to the enhancement of long-range dependencies. This highlights the effectiveness of our hybrid structure in capturing both global and local information. We replace Bi-Mamba with a full-precision Mamba block to evaluate the impact of binarization. Bi-Mamba achieves a 79\% reduction in parameters and an 88\% reduction in total computation costs, with a reasonable performance drop of 0.79dB in PSNR. Specific to the Mamba blocks, as shown in Figure \ref{figure7}, Bi-Mamba significantly compresses the projections, including linear and convolution layers, reducing parameters and computation costs by \textbf{96\%} and \textbf{97\%}.

\vspace{-4mm}
\paragraph{Global Visual Embeddings}
To further explore the global visual embeddings, we analyze the effects of embedding positions across different control matrices, as shown in Table \ref{figure4}. Our proposed global visual embedding method on $\mathbf{B}$ improves PSNR by 0.05dB and SSIM by 0.001, demonstrating that appropriately integrated extra global information can enhance performance. However, applying global visual embeddings to all control matrices or the $\Delta$ matrix results in a slight performance decline. As shown in Formula \ref{eq:9}, $\Delta$ controls both $\overline{\mathbf{B}}$ and $\overline{\mathbf{A}}$, meaning the latent vector $h_k$ is affected in a cumulative product form by global visual information, which reduces its impact and causes $h_k$ to become unstable. In contrast, embedding the global visual vector with $\mathbf{B}$ directly influences the input and impacts the latent vector through cumulative summation, preserving usable information more effectively. Otherwise, embedding into $\mathbf{C}$ avoids cumulative product but has a limited impact on $h_k$, yielding only a 0.01dB improvement in PSNR.

  
    
    
    
    
  
\begin{table}

    \centering
    \fontsize{10}{12}\selectfont

        \centering
        
          \vskip 2mm
          
          \setlength{\tabcolsep}{1.0mm}{
          \begin{tabular}{lcccc}
            
            \toprule
            Modules&Params (M)&OPs (G) & PSNR/SSIM   \\
            
            \midrule
            w/o Mamba& 1.25  & 5.32  & 35.17/0.964       \\ 
            FP Mamba& 6.15  & 54.07  & 37.73/0.971       \\ 
           \textbf{Bi-Mamba (Ours)} & 1.28  & 6.56  & 36.95/0.975       \\ 
            
    \bottomrule
    
  \end{tabular}}
  \caption{Ablation study on the Bi-Mamba block. Bi-Mamba is crucial for maintaining performance, while it significantly reduces both parameters and operations with reasonable performance loss compared with the full precision version.}
    \label{tab:table3}
    \vspace{-4mm}
\end{table}
\begin{table}

    \centering
    \fontsize{10}{12}\selectfont

        \centering
        
          \vskip 2mm
          
          \setlength{\tabcolsep}{1.3mm}{
          \begin{tabular}{lccccc}
            
            \toprule
            SE Position & None& All &$\mathbf{C}$ &$\mathbf{\Delta}$  & \textbf{$\mathbf{B}$ (Ours)}   \\
            
            \midrule
            PSNR &36.90  & 36.89  & 36.91  & 36.83  & \textbf{36.95}      \\ 
            SSIM &0.974  & 0.975  & 0.974  & 0.974  & \textbf{0.975}      \\

    \bottomrule
    
  \end{tabular}}
  \vspace{-2mm}
  \caption{Ablation study on the position of global visual embeddings. Embedding into the control matrix $\mathbf{B}$ enhances global capacity while maintaining a stable influence on the latent vector $h_k$.}
    \label{tab:table4}
    \vspace{-5mm}
\end{table}

%% file: sec/5_conclusion.tex
\vspace{-1mm}
\section{Conclusion}
\vspace{-1mm}
\label{sec:conclusion}
In this paper, we presented a lightweight binarized Mamba-Transformer network (BMTNet) for Quad Bayer HybridEVS demosaicing. First, we presented a binarized global visual encoding branch to acquire additional global information, which effectively enhances Mamba's global capacity. Second, we introduced a binarized Mamba-Transformer structure to reduce the model complexity. The pioneering binarization of Mamba and the fusion of extra global visual embeddings reduce computational complexity by compressing non-essential projections while enhancing performance through precise integration. Experiments conducted on eight diverse datasets demonstrate that our BMTNet outperforms other BNN methods while achieving results comparable to full-precision models at a minimal computational cost. Our approach expands the capabilities of BNNs and offers an efficient and high-performing demosaicing solution for resource-constrained HybridEVS.
\vspace{-4mm}
\paragraph{Acknowledgments:} This work was supported by the National Natural Science Foundation of China (grant No. 62350710797), the Key Program of Technology Research from Shenzhen Science and Technology Innovation Committee under Grant JSGG20220831104402004.